\documentclass{article}

\usepackage[position, final]{neurips_2026}

\usepackage[utf8]{inputenc} %
\usepackage[T1]{fontenc}    %
\usepackage{hyperref}       %
\usepackage{url}            %
\usepackage{booktabs}       %
\usepackage{amsfonts}       %
\usepackage{nicefrac}       %
\usepackage{microtype}      %
\usepackage{xcolor}         %
\usepackage{amsmath}
\usepackage{graphicx}
\usepackage{tikz}
\usetikzlibrary{positioning}
\usepackage{mathtools}
\usepackage{centernot}
\newcommand\independenT[2]{\mathrlap{#1#2}\mkern4mu{#1#2}}
\newcommand\independent{\protect\mathrel{\mathpalette\independenT\perp}}
\newcommand{\indep}{\independent}
\newcommand{\nindep}{\centernot\indep}
\hypersetup{
    colorlinks=true,
    citecolor = {blue!50!gray},
    linkcolor = {red!50!gray},
    urlcolor = {green!50!gray},
    pdfborder={0 0 0} %
}

\title{%
Causal discovery needs explicit epistemic standards
}

\author{%
  Harald Kugler \\
  Department of Mathematical Sciences \\
  University of Copenhagen, Denmark \\
  \texttt{hku@math.ku.dk} \\
  \And
  Leonard Henckel \\
  School of Mathematics and Statistics \\
  University College Dublin, Ireland \\
  \texttt{leonard.henckel@ucd.ie} \\
  \AND
  Sebastian Weichwald \\
  Department of Mathematical Sciences \\
  University of Copenhagen, Denmark \\
  \texttt{sweichwald@math.ku.dk} \\
}

\newtheorem{definition}{Definition}

\newtheorem{example}{Example}

\begin{document}

\maketitle

\begin{abstract}
    There are many causal discovery algorithms,
    but few guiding principles for choosing between them.
    Methods are usually compared through three lenses:
    identifiability, consistency, and benchmarks.
    Each introduces assumptions about the data-generating process that are often untestable.
    We argue that there are no shared standards for comparing, prioritizing, or even discussing those assumptions.
    We support this claim by showing that
    (i)~identifiability assumptions are often preferred by convention rather than by principle, 
    (ii)~uniform consistency requires further untestable assumptions,
    while 
    the weaker pointwise consistency has arguably narrowed algorithm design,
    and 
    (iii)~benchmarks are not neutral tests but carry an epistemic role of their own:
    they quietly fix which regimes matter, which guarantees come into play, and what counts as success.
    In doing so,
    they shape what
    algorithm comparisons are taken to show
    and thus perform some of the same epistemic work as theory,
    but less transparently.
    In this position paper we argue for a more transparent, integrated practice:
    Make the epistemic standards for discussing identifiability and consistency assumptions explicit, especially their failure modes and scope conditions; design benchmarks to operationalize those standards, by stress-testing and ablating both.
    Then, theoretical and empirical results jointly clarify when, why, and how algorithms 
    may work rather than accumulate as disconnected advances.
\end{abstract}
\section{Introduction}

Causal discovery research has produced a large and still growing collection of methods for learning causal graphical structure from data.%
\footnote{
This includes, 
among countless others,
structure learning algorithms
based on observational, interventional, or time-series data \citep{spirtes2001causation,hauser2012imecs,brouillard2020differentiable,mooij2020joint,runge2023timeseries},
targeting directed acyclic graphs or their equivalence classes \citep{chickering1995bayesian}
and ancestral graphs if the observed variables are not causally sufficient \citep{magliacane2016ancestral}.
Approaches range from
constraint-based methods \citep{spirtes1991algorithm}
to score-based ones
based on Bayesian or penalized likelihood principles \citep{cooper1992bayesian,heckerman1995learning},
under distributional assumptions
\citep{shimizu2006linear,peters2014equalvar,park2020identifiability}
or functional restrictions on the causal model 
\citep{buhlmann2014cam,peters2014causal}.
Search strategies include
exhaustive discrete search \citep{koivisto2004exact,silander2006simple},
local discrete search procedures
operating on graphs, equivalence classes, or variable orderings via local moves
\citep{teyssier2012ordering,lam2022greedy,andrews2023fast},
typically implemented greedily \citep{spirtes1995learning,meek1997graphical,chickering2002optimal}
and sometimes employing iterated local search
\citep{wienobst2025embracing},
as well as methods based on
answer set programming \citep{hyttinen2014constraint},
continuous relaxations \citep{aragam2015concave,zheng2018notears},
graph neural networks \citep{yu2019dag},
Bayesian Markov chain Monte Carlo schemes \citep{viinikka2020towards},
generative flow models \citep{deleu2022bayesian},
and score matching \citep{montagna2023scalable}.
Recent surveys include
those by
\citet{glymour2019review,vowels2022d,kitson2023survey}.
}
However, choosing between these methods remains difficult, and recent surveys largely bypass their use in real-world applications altogether \citep{brouillard2025landscape}.
The problem is no longer a lack of assumptions under which recovery is possible,
or a lack of algorithms that succeed under some conditions,
but a lack of guiding principles to navigate the existing landscape of methods.

Methods are usually compared through three lenses:
(i)~identifiability asks what can in principle be recovered,
(ii)~consistency asks if recovery should be expected as sample sizes grow, and
(iii)~benchmarks ask what works in some finite-sample regimes.
Each lens injects implicit or explicit assumptions about the data-generating process. 
However, none comes with a principled, practice-facing community standard for how to justify, compare, or prioritize those assumptions.
As a result, assumptions are studied in isolation, justified mainly by the positive results they enable, and rarely in a picture of when they fail.
Therefore, the landscape of structure learning algorithms, methods, assumptions, guarantees, benchmark settings, trade-offs, and failure modes is hard to read.

\textbf{Identifiability} research shows this symptom clearly.
Some assumptions 
become de facto standards and are debated extensively, while others
are barely discussed.
The choice between them is driven more by convention than by logical hierarchy, testability, or empirical plausibility. 
In Section~\ref{sec:ident}, we make this concrete by contrasting faithfulness with covariance-faithfulness.
From there we draw the conclusion that we can change how we navigate the landscape of assumptions.
Once covariance-faithfulness is taken seriously,
distinctions such as linear versus nonlinear,
Gaussian versus non-Gaussian,
or additive versus non-additive
no longer organize the space in the same way.
Seemingly separate lines of work can then be re-read as living under one assumption instead and algorithms developed for one of these settings turn out to apply across all of them.
Additionally, we examine further assumptions that enable structure recovery yet would likely not be recognized as legitimate under current practice, although they tend to hold in standard benchmark settings.

\textbf{Consistency} raises a parallel concern. 
Pointwise consistency is a weak design criterion
as it does not prevent poor finite-sample behavior.
As we show in Section~\ref{sec:cons},
pointwise consistency is relatively easy to obtain
and yet it has kept algorithm design conservative and based on narrow adaptations,
even though less narrowly developed algorithms can outperform them.
Uniform consistency is a stronger standard
as it provides an asymptotic worst-case guarantee over an entire model class,
but only under additional assumptions.
Because consistency guarantees are relative to an identifiability regime,
analyzing consistency and identifiability in isolation
can obscure what the guarantees amount to,
and is a poor epistemic guide.

\textbf{Benchmarks} are easy to underestimate as playing a merely empirical role, but they carry an epistemic role of their own. 
By fixing the distribution of test graphs, edge densities, functional forms, noise distributions, sample sizes, and evaluation metrics, benchmarks quietly determine which regimes matter, which guarantees come into play, and what counts as success.
In Section~\ref{sec:bench}, we present carefully designed experiments that go beyond familiar benchmark pathologies to demonstrate how benchmarks perform some of the same epistemic work as theory, but less transparently.
This becomes visible when methods outperform competitors while their formal assumptions are violated, or when benchmark success cannot be justified by current theory. 

We claim that these three lenses cannot be studied
in isolation.
To make the core issues arising from the interplay of each lens's implicit epistemic standards precise,
we develop our argument in a deliberately chosen, narrow context:
We assume causal sufficiency with the goal of learning either a directed acyclic graph or its Markov equivalence class.
The issues only become harder in the broader space of causal discovery.
\textbf{Our position is that causal discovery needs to make epistemic standards for comparing assumptions, interpreting guarantees, and designing benchmarks explicit.}
Theory should not merely add another mathematically pleasing assumption to the list, but should be discussed comparatively.
Benchmarks should not merely reward methods that align with one chosen regime but stress-test and systematically ablate those assumptions.
Together, the three threads we develop outline a direction in which future identifiability, consistency, and algorithm work could change what practitioners can do, and in which characterizing failures counts as progress.
\section{Preliminaries}

\subsection{Notation and Definitions}

We follow standard definitions from the causal graphical modeling literature. 
For a comprehensive overview, see, for example, \citet{spirtes2001causation, pearl2009causality, peters2017eleci}. 
Formal statements are collected in Appendix~\ref{app:defs}.

We assume that the underlying causal structure can be represented by a directed acyclic graph (DAG) with
vertex set \(V = \{X_1, \dots, X_p\}\) and 
edge set \(E \subset \{ X_i \rightarrow X_j \mid X_i, X_j \in V \}\). 
The set of parents of a node \(X_j\) is denoted by \(\operatorname{Pa}(j)\), 
and the average (in)degree $\bar{d}$ of a DAG is defined as the average number of parents per node.

DAGs can be used to encode conditional independence statements via a graphical criterion called d-separation \citep{geiger1990identifying}. 
We identify each vertex $X_i$ with a random variable.
Concretely, we say that a distribution $\mathbb{P}$ over $V$
is Markov to a DAG $\mathcal{G}=(V,E)$ if every d-separation in $\mathcal{G}$ \citep{geiger1990identifying} implies the corresponding conditional independence in $\mathbb{P}$.
If both implications hold,
that is,
$X_A \perp_{\mathrm{d\text{-}sep}} X_B \mid X_Z \iff X_A \indep_\mathbb{P} X_B \mid X_Z$
for all disjoint $X_A,X_B,X_Z \subseteq V$,
then $\mathbb{P}$ is faithful to $\mathcal{G}$ (Definition~\ref{def:faithful}).
Two DAGs are Markov-equivalent if every distribution that is Markov to one is also Markov to the other; an equivalence class under this relation is called a Markov equivalence class (MEC) \citep{chickering1995bayesian}.
Each MEC can be represented by a completed partially directed acyclic graph (CPDAG): edges that point in the same direction across all DAGs in the MEC remain directed in the CPDAG, while edges that point in different directions in different DAGs are represented as undirected \citep{chickering2002learning}.

A structural causal model (SCM)
associated to a DAG \((V,E)\)
is a tuple \((V,\allowbreak \{f_j\}_{j=1}^{p},\allowbreak \{N_j\}_{j=1}^{p})\), 
where each variable \(X_j\) satisfies a structural equation of the form
\(
X_j = f_j(X_{\operatorname{Pa}(j)}, N_j),
\)
where \(X_{\operatorname{Pa}(j)}\) are the
direct causes of $X_j$, \(N_j\) an exogenous noise variable independent of all other noise terms,
and \(f_j\) a measurable function.
Throughout, we assume structural minimality of the SCM
such that every parent in the graph is relevant to the corresponding child node's
observational, interventional, or counterfactual distribution;
this is a convention on representation that avoids redundant SCM representations
and is not a substantive assumption about the data-generating process, see Remark~6.6 in \citet{peters2017eleci} for more details.

An additive noise model (ANM) is an SCM with each structural equation decomposing additively as
\(
    X_j = g_j(X_{\operatorname{Pa}(j)}) + N_j,
\)
where $g_j$ is a measurable function. 
If each $g_j$ in an ANM is linear, we call it a linear ANM.

\subsection{Algorithms}
We focus on a representative, tractable corner of the causal discovery landscape and consider algorithms designed for linear ANMs under causal sufficiency. 
Beyond this shared setting, the algorithms differ in their further assumptions and guarantees.

\textbf{DirectLiNGAM} \citep{shimizu2011directlingam} finds a DAG 
via
first, finding a causal ordering by iteratively selecting an exogenous variable based on independence from remaining variables' regression residuals, and 
second, selecting each variable's parents in that order using adaptive Lasso regression.

\textbf{VarSortRegress} \citep{reisach2021beware} exploits the observation that marginal variance tends to increase along the causal order in commonly used benchmarks to recover a causal order. It then estimates the DAG by regressing each variable on its predecessors in this order, using Lasso regression 
with the Bayesian Information Criterion (BIC) \citep{schwarz1978bic} for model selection.

\textbf{FLOP} \citep{wienobst2025embracing} is a score-based causal discovery algorithm originally introduced for linear Gaussian ANMs that optimizes the BIC score via discrete search over topological orders and parent sets.
It returns a CPDAG. 
Since we use FLOP as a DAG-learner, we consider a random DAG from the MEC encoded by its CPDAG output. 

\subsection{Evaluation Metrics}
It is a common observation that the choice of metric shapes relative algorithm performance (see also Figure~\ref{fig:metric}).
Related work on identifiable representation learning shows more broadly that evaluation metrics can encode additional assumptions about the data-generating process, the estimate, and which aspects of the theoretical recovery target count as successfully recovered \citep{joshi2026whoguards}.
We therefore
use two metrics that capture different aspects of the discovered DAG's structural accuracy.
In the main text, we present the structural Hamming distance (SHD) \citep{tsamardinos2006shd}, which quantifies the total number of edge additions, deletions, and re-orientations required to transform the estimated graph into the true DAG.
For each experiment, we present analogous results using the ancestor adjustment identification distance (AID) \citep{henckel2024aid} in Appendix~\ref{app:additionalfigures}.
The ancestor AID
quantifies the accuracy of the learned graph by counting all pairs of single treatment and outcome nodes
where the ancestors of the treatment in the learned graph
do not form a valid adjustment set (for the effect of the treatment onto the outcome node) in the true graph.
It therefore is a task-oriented metric that evaluates how well the learned graph can be used for the downstream task of causal effect identification via ancestor adjustment.
Ancestor AID evaluation emphasizes the found ordering and
if the found graph respects the causal orders of the true graph
the ancestor AID is zero.

\section{Identifiability:
Which assumptions count as fundamental?}\label{sec:ident}

The causal discovery literature has produced a rich variety of identifiability assumptions characterizing 
when graphs can be learned from observational distributions. 
In this section, we argue that these assumptions are often chosen by implicit and inconsistent standards. 
There is no principled criterion by which one assumption is preferred over another. 
We illustrate this through two observations:
First, well-established assumptions such as faithfulness and covariance-faithfulness form no logical hierarchy,
and second, the community applies an inconsistent standard in deciding which assumptions are acceptable.
Together, these suggest that the identifiability literature, while valuable for mapping the space of what is recoverable under which assumptions,
has no clear sense of direction unless the community agrees on the epistemic standards to decide between assumptions.

The Markov and faithfulness assumptions (Definition~\ref{def:faithful}) are treated as the fundamental baseline assumptions in the literature \citep{spirtes2001causation, chickering2002optimal}.
Under causal sufficiency they identify the MEC of a causal DAG
and both have been debated philosophically \citep{cartwright2001bayesnets}.
On one hand, the Markov assumption holds by construction: 
once one commits to  the
data-generating process over the chosen variables to admit an SCM-based causal description,
and hence to learning a causal graph over these variables as a meaningful task
(itself a substantive representational commitment
\citep{dawid2010beware,eberhardt2016green,rubenstein2017causal,scholkopf2021toward,jorgensen2025causal}),
the induced distribution is Markov to the causal graph (Proposition 6.31 in \citealp{peters2017eleci}).
Faithfulness, on the other hand, may be violated. 
It does, however, hold generically for many parametric models \citep{spirtes2001causation, boeken2024nonparametric}. 
Nonetheless, its plausibility in real-world settings has been debated at times, as it is, for example, naturally violated in settings with mechanistic relationships \citep{triantafillou2017predicting}.
In the following, we argue that its standard treatment as a fundamental baseline is also not justified from another perspective: it does not logically dominate many other arguably equally plausible identifiability assumptions.

For example, covariance-faithfulness is an equally valid but distinct condition for MEC identifiability.
Intuitively, it requires that whenever two variables show no (partial) correlation in the data, the graph must reflect this through a corresponding d-separation.
Formally, a distribution $\mathbb{P}$ over $V = \{X_1, \dots, X_p\}$ is covariance-faithful to $\mathcal{G}$ if, for all distinct $X_i, X_j \in V$ and all $X_S \subseteq V \setminus \{X_i, X_j\}$, we have
$\rho(X_i, X_j \mid X_S) = 0 \iff X_i \perp_{\mathrm{d\text{-}sep}} X_j \mid X_S$,
where $\rho(X_i, X_j \mid X_S)$ denotes the (linear) partial correlation, defined as the correlation between the residuals obtained from linearly regressing $X_i$ and $X_j$ on $X_S$ (Definition~\ref{def:cov-faithful}).
The term covariance‑faithfulness reflects that verifying it with respect to a given graph requires only second‑order information, namely (inverses of) (submatrices of) the covariance matrix.

In the Gaussian case, covariance-faithfulness coincides with faithfulness; outside the Gaussian setting it should be understood as a restriction to second‑order information.
In general, although equivalent in the linear Gaussian case, neither zero (partial) correlation nor conditional independence implies the other.
Through the following two counter-examples,
we exhibit two SCMs that make this concrete,
showing that $\mathbb{P}$ being 
faithful to $\mathcal{G}$
neither implies nor is implied by
$\mathbb{P}$ being
covariance-faithful to $\mathcal{G}$.
In both cases, causal discovery relying on the respective other assumption will yield a wrong graph (that is, not the underlying causal graph but one that exactly encodes the vanishing correlation or independence constraints).

\begin{example}[Covariance-faithfulness does not imply faithfulness]\label{ex:one}
    Consider the SCM
    \begin{align*}
        X_1 &= N_1 \\
        X_2 &= -3 + 3 \cdot \mathbf{1}\{X_1=2\} + N_2 \\
        X_3 &= -3 + \mathbf{1}\{X_1 > 0\} + \operatorname{sign}(X_2) \cdot N_3
    \end{align*}
    with $N_1\sim \mathrm{Unif}(\{0,1,2\})$, $N_2,N_3\sim \mathrm{Unif}(\{-1,1\})$ jointly independent.
    The induced distribution $\mathbb{P}$ is uniform over $12$ configurations, and the causal graph $\mathcal{G}_0$ is the fully connected triangle
    \begin{tikzpicture}[baseline=(B.base), overlay=false]
    \node (A) {$X_1$};
    \node (B) [right=0.9em of A] {$X_2$};
    \node (C) [right=0.9em of B] {$X_3$};
    \draw[->, line width=0.3pt] (A) -- (B);
    \draw[->, line width=0.3pt] (B) -- (C);
    \draw[->, bend left=18, line width=0.3pt] (A) to (C);
    \end{tikzpicture}\!.
    All entries of the covariance matrix $\Sigma$ and precision matrix $\Sigma^{-1}$ are nonzero, so $\mathbb{P}$ is covariance-faithful to $\mathcal{G}_0$.
    Yet $X_2 \indep X_3 \mid X_1$ (the only conditional independence among $X_1,X_2,X_3$), whereas under faithfulness $\mathcal{G}_0$ would imply
    $X_2 \nindep X_3 \mid X_1$ via the edge $X_2 \to X_3$. Hence $\mathbb{P}$ is not faithful to $\mathcal{G}_0$. See Appendix~\ref{app:examples} for the enumeration, the matrices $\Sigma,\Sigma^{-1}$, and the derivation.
\end{example}

\begin{example}[Faithfulness does not imply covariance-faithfulness]\label{ex:two}
    Consider the SCM
    \begin{align*}
        X_1 &= N_1 \\
        X_2 &= X_1 + N_2 \\
        X_3 &= 2X_1 - X_2 + N_3
    \end{align*}
    with $N_1, N_2, N_3$ jointly independent, zero mean, unit variance, and non-Gaussian with finite fourth moment; the reduced form is $X_2 = N_1 + N_2$, $X_3 = N_1 - N_2 + N_3$. The causal graph $\mathcal{G}_0$ is the fully connected triangle
    \begin{tikzpicture}[baseline=(B.base), overlay=false]
    \node (A) {$X_1$};
    \node (B) [right=0.9em of A] {$X_2$};
    \node (C) [right=0.9em of B] {$X_3$};
    \draw[->, line width=0.3pt] (A) -- (B);
    \draw[->, line width=0.3pt] (B) -- (C);
    \draw[->, bend left=18, line width=0.3pt] (A) to (C);
    \end{tikzpicture}\!.
    Direct computation gives $\operatorname{Cov}(X_2, X_3)=0$ while $X_2,X_3$ are adjacent in $\mathcal{G}_0$, so covariance-faithfulness fails.
    Yet $\mathbb{P}$ is faithful to $\mathcal{G}_0$: no conditional independence holds among $X_1,X_2,X_3$. In particular, despite $\operatorname{Cov}(X_2, X_3)=0$, the Darmois--Skitovich theorem gives $X_2 \nindep X_3$ because $N_1,N_2$ are non-Gaussian. See Appendix~\ref{app:examples} for covariance matrix $\Sigma$, precision matrix $\Sigma^{-1}$ and the verification of the remaining dependences.
\end{example}

The above examples show that faithfulness and covariance-faithfulness are logically independent:
neither implies the other. 
Both are equally untestable, 
yet faithfulness has become the backbone of much of the causal discovery literature, while covariance-faithfulness has received comparatively little attention \citep{spirtes98covfaith, buehlmann2010pcsimple, schultheiss2023gaussianlikelihood}. 

To our knowledge, covariance-faithfulness has only been studied in linear ANMs. 
But as a second-order condition, it can hold under nonlinear functional forms or non-additive errors.%
\footnote{
    For example, it holds in the SCM $X_1 \to X_2 \to X_3$ with $X_1 = N_1$, $X_2 = X_1^3 + N_2$ and $X_3 = X_2(1 + N_3)$, where $N_1, N_2, N_3$ are independent standard normal. Although the mechanism of $X_2$ is nonlinear and the noise of $X_3$ is multiplicative, the partial correlation of $X_1$ and $X_3$ given $X_2$ vanishes because $N_3$ has mean zero, and all d-connected pairs remain (partially) correlated because $X_1^3$ has a nonzero linear component. Covariance-faithfulness can, however, often fail. See Example 7.9 in \citet{peters2017eleci}.
}
It can be stated directly as a distribution-level assumption on the generated distribution, rather than an implication of an assumption-restricted SCM class. Assuming covariance-faithfulness, the MEC is identifiable as the unique CPDAG that encodes those and only those partial correlations that are vanishing, which can be read off the covariance matrix, and does not require commitment to functional or distributional forms of an underlying SCM.
The algorithm FLOP, for example, was introduced only for the linear, additive Gaussian case under faithfulness (and causal sufficiency) \citep{wienobst2025embracing}.
It could, however, have been formulated to handle linear and nonlinear, Gaussian and non-Gaussian, additive and non-additive SCMs by assuming only covariance-faithfulness. 
In fact, \citet{wienobst2025embracing} themselves present ablation studies that show no substantial performance drop when deviating from the stated Gaussian noise assumption, 
which could be explained by these settings still satisfying covariance-faithfulness.

Why this generalization has not been pursued is unclear.
One might suspect that faithfulness enjoys stronger theoretical support than covariance-faithfulness, but this is not the case.
The traditional zero-measure argument \citep{spirtes1993causation,meek1995faithful} was formulated for linear Gaussian SCMs, and in that setting faithfulness and covariance-faithfulness coincide.
The classical justification applies therefore to both equally.
We see no principled reason to prefer faithfulness combined with functional or distributional restrictions over covariance-faithfulness alone.
The observed community preference must therefore rest on standards that are implicit.

The above illustrates a general problem in the causal discovery literature:
We have no principled means to decide between identifiability assumptions other than mathematical beauty.
This applies equally to stronger assumptions that allow full identifiability of the DAG, where various functional and distributional assumptions have emerged.
For example,
\citet{shimizu2006linear} establish identifiability under linear non-Gaussian ANMs. 
\citet{peters2014equalvar} show that the full DAG is recoverable when Gaussian noise terms have equal variance. \citet{park2020identifiability} establish a more general condition exploiting marginal and conditional variances with heterogeneous error variances.

\begin{figure}
    \centering
    \includegraphics[width=\linewidth]{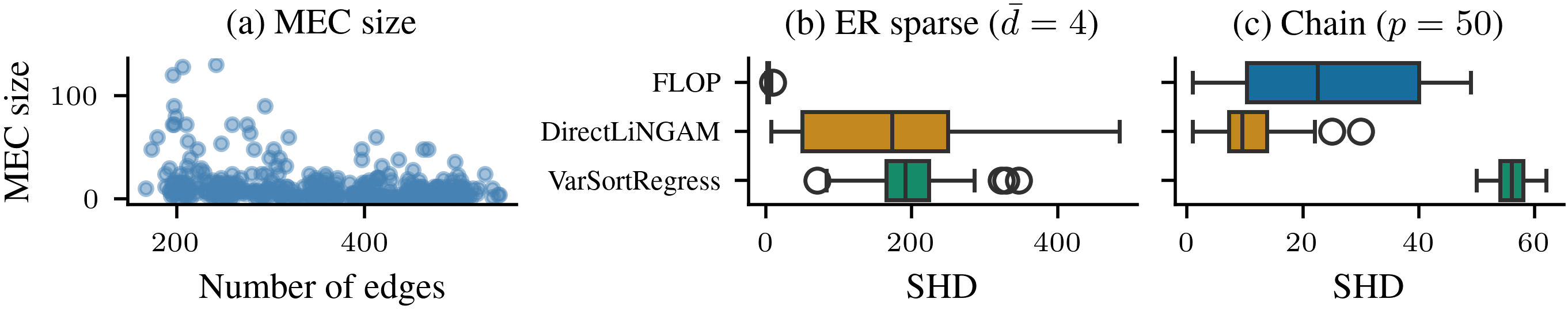}
    \caption{
    FLOP vs.\ DirectLiNGAM vs.\ VarSortRegress on linear ANMs with $p=50$ nodes, $n=1000$ samples, and generalized normal noise ($\beta=1.4$); 50 sampled graphs per setting.
    \textit{(a)}~MEC size vs.\ number of edges on ER graphs (one point per graph): small across the considered range.
    \textit{(b)}~SHD to the true DAG on ER graphs: although FLOP targets the MEC rather than the DAG, the small MEC leaves few edges undirected, and FLOP outperforms DirectLiNGAM despite the latter's assumptions being satisfied.
    \textit{(c)}~The same comparison on chains without colliders (MEC size of $50$): FLOP recovers the skeleton (SHD $\leq 49$) but orients edges at random, and is therefore outperformed by DirectLiNGAM. 
    See Figure~\ref{fig:graphtype-aid} for the corresponding ancestor AID results.
    }
    \label{fig:MEC+type}
\end{figure}

It is relatively easy to come up with alternatives allowing DAG identifiability, as the following two examples illustrate. 
First, we could impose the structural assumption that the MEC has size one, that is, the true DAG is the unique member of its equivalence class.
This may seem strong, given that MECs are known to potentially be large \citep{jahn2025lowerboundssizemarkov}.
Yet, for Erdős–Rényi (ER) graphs \citep{erdos1960evolution}, commonly used in benchmarks,
this assumption is far from unrealistic as their MEC sizes are often very small (Figure~\ref{fig:MEC+type}(a)).
This has direct consequences for algorithm performance, as shown in Figure~\ref{fig:MEC+type}(b) and \ref{fig:MEC+type}(c), and discussed further in Section~\ref{sec:bench}.

Second, we could reframe sortability (Definition~\ref{def:sortability}) as an identifiability assumption.
Varsortability \citep{reisach2021beware} measures the extent to which the causal order of the variables agrees with the order induced by their marginal variances. When it equals one, the two orders coincide, so a causal order can be recovered by sorting variables by marginal variance, and the true DAG by sparse regression \citep{shojaie2010sparseregression}.
The same applies to $R^2$-sortability \citep{reisach2023scale}, where variables are sorted by the fraction of their variance explained by the remaining variables.
A third instance is the assumption that the temporal order is known. 
\citet{reisach2021beware} show that varsortability is high in commonly used benchmarks and that algorithms such as NOTEARS \citep{zheng2018notears} implicitly exploit this property. 
This has led to the development of benchmarks specifically designed to eliminate sortability patterns \citep{andrews2024better, hermann2025uumcscm, ormaniec25iscm}, therefore treating varsortability as an artifact to be removed rather than an identifiability assumption.
However, there is empirical evidence that sortability may reflect genuine properties of some real-world data \citep{lohse2025sortability}.
Further, one may argue that knowledge of the temporal order is realistic in experimental setups where variables are measured at known time points, such as longitudinal studies or time-series experiments \citep{bang2026improving, reisach2026casetimecausaldags}, and that the widespread framing of the causal discovery task without such knowledge is in fact itself contrived.

While the established assumptions undoubtedly provide useful mathematical insights and explore the theoretical limitations of identifiability, 
it remains unclear what makes them more plausible than alternatives like those above, which, to our knowledge, have not been considered before.
Common benchmark instances are concentrated near the boundary of these two exemplary identifiability assumptions (that is, MEC size and varsortability are often close to one), and for varsortability, there is even real-world evidence. 
Why then embrace conceptually more restrictive assumptions while dismissing simpler ones like a MEC size of one? 
Why spend effort removing sortability from benchmarks rather than exploiting it directly?
These questions deserve an open discussion, so that the standards for what constitutes a valid identifiability assumption do not remain implicit. 
Specifically, this means that we should discuss also when an assumption may not hold, not only when it may hold, otherwise new results accumulate as disconnected advances.

\section{Consistency: Which guarantees count as meaningful?}\label{sec:cons}

Here, we discuss standards around statistical guarantees and their epistemic role in the causal discovery literature. 
It is natural to study which statistical guarantees causal discovery algorithms possess as estimators.
Although well-known to be too weak to ensure satisfactory performance,
pointwise consistency is a common property to require of an estimator.
For causal discovery algorithms,
it is often the only statistical guarantee
\citep{chickering2002optimal,nazaret2025extremely}.
We argue that pointwise consistency has failed in its epistemic role of incentivizing the development of good causal discovery algorithms.
First, most causal discovery algorithms can easily be used in ways that ensure pointwise consistency.
Second, proofs are easier for simple and theoretically tractable algorithms \citep{spirtes2001causation,colombo2012learning,nandy2018high} even though these often have mediocre performance \citep{rios2021benchpress}.

We illustrate both points in score-based causal discovery. 
Here, the BIC score has long been known to be pointwise consistent for graph recovery \citep{koller2009probabilistic}. 
Exact algorithms that find a graph with globally optimal BIC score \citep{koivisto2004exact,silander2006simple}
are therefore pointwise consistent graph estimators,
simply because the underlying score is.
Unfortunately,
exhaustive search over the graph space 
has exponential runtime and is only feasible up to about $30$ nodes.
To resolve this runtime bottleneck, the development of greedy equivalence search (GES) was an important milestone:
it builds on the observation that the BIC score is also locally consistent, which guarantees recovery of the MEC via a greedy search that first iteratively adds and then deletes the edge with the greatest score improvement \citep{chickering2002optimal}.

Pointwise consistency of GES does not, however, guarantee good finite-sample performance, since greedy search can get stuck in local optima and fail to find a globally BIC-optimal graph.
Follow-up algorithms such as XGES \citep{nazaret2025extremely} are motivated by this limitation.
Yet, to preserve consistency via the GES proof strategy, XGES retains greedy insertion and deletion phases, with the new heuristic acting only as an outer wrapper reinitializing the search \citep{nazaret2025extremely}.
Therefore, the reliance on the GES proof template constrains how far the algorithm can depart from GES.
This is unnecessary. 
Given a pointwise consistent baseline like GES, any procedure producing a candidate graph
can be made consistent by running GES in parallel and returning whichever output attains the better score.
This separates score-level guarantees from the convergence behavior of the candidate-generating method and does not even
require it 
to 
be score-based.
New algorithms are therefore not bound to greedily optimize the BIC score.
For example, BOSS \citep{andrews2023fast} and FLOP \citep{wienobst2025embracing} already embody this philosophy, optimizing the Gaussian BIC through broader discrete search techniques, and deliver substantial performance gains over GES and alternatives.
Decoupling algorithm design from consistency proofs
is particularly valuable for continuous approaches such as DAGMA \citep{bello2022dagma},
which optimize a relaxed objective and then threshold and project back to a discrete graph,
making direct consistency proofs technically challenging.

By contrast, uniform consistency provides a stronger, practically relevant epistemic standard than pointwise consistency: 
It ensures convergence to the ground-truth graph at a uniform rate over all models.
This means there is no sequence of particularly challenging models for which convergence fails in arbitrarily large samples.
However, \citet{robins2003uniform} show that uniform consistency is unachievable under faithfulness alone, so additional assumptions are necessary. 

One such assumption is $\lambda$-strong faithfulness (Definition~\ref{def:strong-faithful}) \citep{zhang2012strong}. 
By requiring that all partial correlations corresponding to d-connections exceed a fixed threshold $\lambda$ (which mirrors conditions for PC's uniform consistency \citep{kalisch2007pcconsistent}), this assumption rules out the near-unfaithful distributions that obstruct uniform convergence \citep{zhang2012strong}.
This comes at a cost.
\citet{uhler2013geometry} show that, for linear Gaussian models, the set of non-$\lambda$-strong-faithful distributions has strictly positive Lebesgue measure and can be surprisingly large. 
Furthermore, its volume grows rapidly with both the number of nodes and the graph density \citep{uhler2013geometry}.
So, $\lambda$-strong faithfulness cannot be 
justified by appealing to its violations forming a zero-measure set,
unlike faithfulness \citep{spirtes1993causation,meek1995faithful}.

To summarize, neither guarantee on offer is unproblematic.
Pointwise consistency is weak and as a design target has shaped algorithm design in ways that are neither obviously helpful nor necessary.
Uniform consistency requires assumptions that are difficult to test.
Nevertheless, consistency-relevant conditions
may silently shape benchmark performance (Section~\ref{sec:bench}).
We therefore argue for a broader discussion of consistency as epistemic standard.
Assumptions need to be explicit and transparent to avoid misattributing differences between algorithms solely to identifiability assumptions instead.
While the literature treats identifiability and consistency as largely separate questions, with advances often in separate papers,
the two are coupled and should always be considered jointly.

\section{Benchmarks:
Which regimes count as informative and representative?
}
\label{sec:bench}

\begin{figure}
    \centering
    \includegraphics[width=\linewidth]{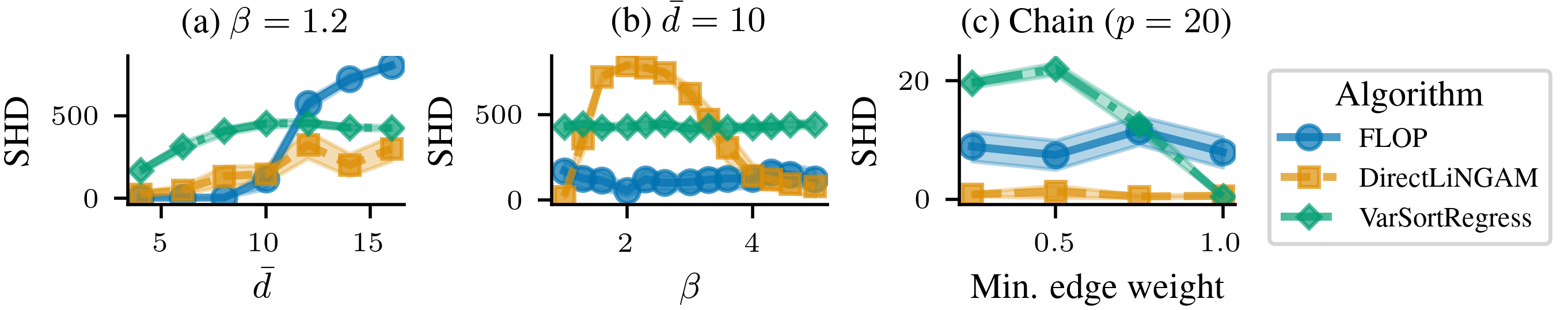}
    \caption{
    FLOP vs.\ DirectLiNGAM vs.\ VarSortRegress on linear ANMs with $n=1000$ samples and generalized normal noise ($\beta$ controls peakedness and tails).
    \textit{(a)}~Varying average degree $\bar{d}$ on ER graphs ($p=50$, $\beta=1.2$): FLOP's performance degrades with density, while DirectLiNGAM and VarSortRegress remain stable.
    \textit{(b)}~Varying the noise distribution parameter $\beta$ on ER graphs ($p=50$, $\bar{d}=10$): DirectLiNGAM is sensitive to $\beta$ and fails close to $\beta=2$ (Gaussian noise, violating its assumptions); FLOP and VarSortRegress are stable.
    \textit{(c)}~Varying edge weights on chains ($p=20$, $\beta=1.2$): VarSortRegress improves with larger absolute weights; FLOP and DirectLiNGAM are stable.
    See Figure~\ref{fig:degradationaid} for the corresponding ancestor AID results.
    }
    \label{fig:axis}
\end{figure}

Causal discovery algorithms are predominantly evaluated on simulated benchmarks, largely because reliable real-world ground truth is scarce \citep{gentzel2019case, brouillard2025landscape}.
However, benchmark performance is known to depend on hidden features of the benchmark setup itself.
Commonly used benchmark instances exhibit high varsortability, or, after standardization, high $R^2$-sortability \citep{reisach2021beware,reisach2023scale}. This could explain why NOTEARS \citep{zheng2018notears} achieves strong DAG recovery performance even though the invoked identifiability theory only guarantees identification of the MEC.
In the following, we work out examples connected to the previous sections:
These examples include identifiability assumptions implicit in the benchmark design, whether the test setting satisfies uniform consistency conditions, and the degree to which algorithm assumptions are met or violated.
Left unexamined, these factors render algorithm comparisons largely uninformative.

As discussed in Section~\ref{sec:ident}, ER graphs typically exhibit small MECs (Figure~\ref{fig:MEC+type}(a)). 
The smaller the MEC, the fewer edges remain unoriented, and therefore recovering the MEC already determines most of the DAG. 
In the limiting case of a size-one MEC the two coincide rendering the DAG identifiable.
This makes algorithms such as exact search for BIC-optimal graphs \citep{koivisto2004exact,silander2006simple}, BOSS \citep{andrews2023fast}, or FLOP \citep{wienobst2025embracing} a suitable choice for DAG recovery, 
even though they are designed only to recover the MEC under their respective assumptions.
Note that a MEC of size one as a DAG identifiability assumption (Section~\ref{sec:ident}) is generally not satisfied in this setting.
Nevertheless, FLOP followed by randomly drawing a DAG from the found small MEC can outperform DAG-learners whose assumptions are met, because the residual orientation ambiguity within a small MEC is minor.
To illustrate this, consider ER graphs with an average degree of four and 1000 samples, where the noise is drawn from a generalized normal distribution with $\beta=1.4$. 
Therefore, the noise is non-Gaussian and the identifiability assumptions of DirectLiNGAM \citep{shimizu2011directlingam} are satisfied.
Despite that, FLOP exhibits superior performance in DAG recovery: 
FLOP recovers the structure robustly and leaves only the few edges of the small MEC to a random draw. When the MEC is small, the errors introduced by random orientation are smaller than the errors of DirectLiNGAM's estimation (Figure~\ref{fig:MEC+type}(b) and \citet{reisach2026topologicalsortingcriterionrandom}).

The picture flips for chains, that is, DAGs whose skeleton is a single path $X_1 - X_2 - \dots - X_p$ and without colliders (their MEC then has size $p$).
In an experiment with these graphs and the other experiment parameters unchanged, FLOP yields an SHD of at most 49, indicating that it recovers the true skeleton. 
However, since the edges are randomly oriented, we see worse performance than for DirectLiNGAM (Figure~\ref{fig:MEC+type}(c)).
So, FLOP's strong DAG recovery performance on ER graphs is incidental to the design of the benchmark.
It works there because the MECs happen to be small.
This underscores the importance of comparisons against algorithms whose assumptions are not fulfilled, as these may still perform well on a given benchmark. 
Recent work has begun to examine algorithm behavior under assumption violations \citep{montagna2023assumption, yi2025the}.

Of course, FLOP's performance is not guaranteed even when the MEC is small. 
As average degree increases, FLOP degrades rapidly while DirectLiNGAM remains comparatively robust (Figure~\ref{fig:axis}(a)). 
The failure is not only one of search but also of score reliability: BIC's optima no longer correspond to the true graph in this regime (Figure~\ref{fig:bic}).
Intuitively, this observation can be understood through $\lambda$-strong faithfulness (Definition~\ref{def:strong-faithful}): Larger parent sets tend to reduce the magnitude of individual partial correlations, pushing the sampled distributions closer to the non-$\lambda$-strong-faithful regime. 
While \citet{uhler2013geometry} establish this theoretically for linear Gaussian models, \citet{olko2025since} demonstrate
empirically that the proportion of $\lambda$-strong-faithful distributions decreases with graph density in nonlinear settings, suggesting the phenomenon is not specific to the linear Gaussian case. 
These observations suggest that benchmark evaluations should not only discuss the identifiability assumptions, but also whether the chosen parameters place the settings within a regime that supports uniform consistency.

DirectLiNGAM has an analogous axis of degradation. 
By manipulating the $\beta$-parameter of a generalized normal distribution, we vary the degree of Gaussianity of the noise. 
DirectLiNGAM's performance drops as the distribution approaches the Gaussian, while FLOP is unaffected by that change: it relies only on second-order information not the distribution (Figure~\ref{fig:axis}(b)).
To our knowledge, no theory has been developed to quantify the relationship between the degree of non-Gaussianity and the sample size DirectLiNGAM needs to exploit it.
Therefore, systematic, gradual ablation along different axes trying to break a method should become best practice.
Mapping out failure modes helps advance the field,
especially in settings where there are no uniform consistency results yet.

More generally, for any algorithm 
we can find
some subspace of the model space where it performs well. 
Figure~\ref{fig:axis}(c) shows that increasing the lower bound of the edge weight interval makes the sampled models more varsortable, and VarSortRegress achieves correspondingly lower SHDs.
Whether the properties of that subspace are framed as artifacts or as implicit theoretical assumptions is a choice, and the current benchmark development largely treats them as artifacts to remove \citep{andrews2024better, hermann2025uumcscm}.
\citet{ormaniec25iscm}, for instance, reduce varsortability through variance standardization (iSCMs); but the same operation restricts the implied edge weights to a smaller region of the parameter space (Figure~\ref{fig:iscm}), where the sampled models lie closer to the hypersurfaces of non-faithful distributions and are more likely to violate $\lambda$-strong faithfulness assumptions. Eliminating one artifact installs another, and we conjecture that no exhaustive benchmark design can escape this:
Any restriction of the model space leaves regions where different algorithms perform better.

Benchmarks are not neutral arbiters of empirical performance. 
By restricting which models, noise distributions, and parameter regions are sampled, they implicitly answer theoretical questions the field has not yet agreed on how to answer explicitly. 
In this sense, benchmarks are already doing the work of theory but less transparently. 
Just as we argued for identifiability and consistency in Sections~\ref{sec:ident} and \ref{sec:cons}, this work should be surfaced and made discussable.
Concretely, this means that algorithm development should report failure modes rather than only successes, and that benchmark choices should be presented as assumptions on the data-generating process, which a practitioner can interrogate, contest, or accept depending on the application.

\section{Conclusion}

Across identifiability, consistency, and benchmarking, the same pattern recurs.
Each lens introduces conditions on the data-generating process, and each is governed by standards that operate implicitly:
which assumptions count as fundamental, which guarantees count as meaningful, which benchmarks count as informative and representative.
Methods accumulate, theoretical frameworks proliferate, and benchmark suites multiply, but the criteria by which they are compared often remain unstated.

We take the position that this is a central bottleneck in causal discovery.
The field does not lack methods, results, or benchmarks; it lacks shared standards for navigating the ones it already has.
The three lenses cannot be addressed separately either, because none of them supplies an epistemic standard on its own: a consistency guarantee is stated relative to an identifiability regime, and both become visible only in the finite-sample regimes and metrics a benchmark happens to select.
Bayesian model selection and posterior uncertainty can operationalize such standards once model spaces, priors, likelihoods, and targets are specified, but they do not determine which assumptions or benchmark regimes are appropriate in the first place.
Causal discovery needs to make these standards explicit.

Several concrete questions follow.
When an identifiability assumption is only approximately satisfied, does performance degrade gradually or fail abruptly, and can finite-sample analyses and benchmark ablations characterize where that change occurs?
This has been studied for faithfulness \citep{uhler2013geometry}, but not systematically across identifiability assumptions.
Whether small MEC size or sortability is a legitimate assumption or a benchmark artifact likewise cannot be decided by theory or benchmark construction alone;
it depends on the downstream task and on whether the metric rewards the relevant recovery target, as the reversals across graph families, noise regimes, and metrics in Section~\ref{sec:bench} illustrate.
More generally, how do benchmark artifacts arise from choices of model classes, parameter regions, and metrics?
We conjecture they cannot be eliminated, only relocated; if so, the question is not which benchmark is neutral but which commitments a benchmark should be designed to make legible.
Several recent studies pursue these questions \citep{weichwald2020causal,reisach2021beware,reisach2023scale,reisach2026topologicalsortingcriterionrandom,montagna2023shortcutscausaldiscoverynonlinear,andrews2024better,hermann2025uumcscm,ormaniec25iscm,lohse2025sortability}.

Answering them requires benchmarks that vary assumption satisfaction gradually, reveal where methods cease to outperform alternatives, and relate these crossovers to finite-sample behavior and identifiability guarantees.
Explicit standards of this kind also let a practitioner ask which assumptions a method requires and whether their application plausibly meets them.
Real-world and physically grounded benchmarks \citep{runge2019InferringCausationTime,goebler2024causalassembly,gamella2025causalchambers} can confront such plausibility judgments with reality.

The change we advocate is therefore modest in form and substantial in consequence.
Treat alternative assumptions as candidates rather than non-standard curiosities.
Treat consistency results as tools for understanding regimes rather than as design constraints.
Treat benchmarks as instruments that encode theoretical commitments and should be designed to surface them.
We have worked through these moves in a deliberately narrow setting; they only become more important in the broader landscape, where the relevant assumptions and targets differ across latent-variable, temporal, interventional, representation-learning, and application-specific settings.
Characterizing where and why algorithms fail should count as progress, not as an inconvenience to be avoided.
Without that shift, further methodological work will produce more results without producing more clarity.

\begin{ack}
    We thank Hubert Drazkowski, Luigi Gresele, Francesco Freni, Silvan Vollmer and Emanuele Marconato for the helpful discussions, insights and feedback on earlier drafts of this work.
    We also express our gratitude to the anonymous reviewers for the feedback during the rebuttal phase.
    HK and SW were supported by a research grant (0069071) from Novo Nordisk Fonden. This research was supported by the Pioneer Centre for AI, DNRF grant number P1.
\end{ack}

\bibliographystyle{abbrvnat}
\bibliography{references}

\newpage
\appendix

\section{Formal Definitions}\label{app:defs}

In the following, we state the formal definitions of relevant concepts used above.

\begin{definition}[Markov property and faithfulness]\label{def:faithful}
    Let $\mathbb{P}$ be a distribution over $V = \{X_1, \dots, X_p\}$ and let $\mathcal{G} = (V, E)$ be a DAG.
    We say that $\mathbb{P}$ is \emph{Markov} to $\mathcal{G}$ if every d-separation in $\mathcal{G}$ implies the corresponding conditional independence in $\mathbb{P}$,
    \begin{align*}
        X_A \perp_{\mathrm{d\text{-}sep}} X_B \mid X_Z \;\implies\; X_A \indep_{\mathbb{P}} X_B \mid X_Z
        \qquad \text{for all disjoint } X_A, X_B, X_Z \subseteq V,
    \end{align*}
    and \emph{faithful} to $\mathcal{G}$ if additionally the converse holds, that is 
    \begin{align*}
        X_A \perp_{\mathrm{d\text{-}sep}} X_B \mid X_Z \;\iff\; X_A \indep_{\mathbb{P}} X_B \mid X_Z
        \qquad \text{for all disjoint } X_A, X_B, X_Z \subseteq V.
    \end{align*}
\end{definition}

\begin{definition}[Covariance-faithfulness]\label{def:cov-faithful}
    Let $\mathbb{P}$ be a distribution over random variables $V = \{X_1, \dots, X_p\}$ and let $\mathcal{G} = (V, E)$ be a DAG. We say that $\mathbb{P}$ is \emph{covariance-faithful} to $\mathcal{G}$ if, for all distinct $X_i, X_j \in V$ and all $X_S \subseteq V \setminus \{X_i, X_j\}$,
    \begin{align*}
        \rho(X_i, X_j \mid X_S) = 0 \;\iff\; X_i \perp_{\mathrm{d\text{-}sep}} X_j \mid X_S,
    \end{align*}
    where $\rho(X_i, X_j \mid X_S)$ denotes the linear partial correlation, that is, the correlation between the residuals obtained from linearly regressing $X_i$ and $X_j$ on $X_S$.
\end{definition}

\begin{definition}[$\lambda$-strong faithfulness]\label{def:strong-faithful}
    Fix $\lambda \in (0,1)$. Let $\mathbb{P}$ be a distribution over $V = \{X_1, \dots, X_p\}$ and let $\mathcal{G} = (V, E)$ be a DAG. We say that $\mathbb{P}$ is \emph{$\lambda$-strong-faithful} to $\mathcal{G}$ if, for all distinct $X_i, X_j \in V$ and all $X_S \subseteq V \setminus \{X_i, X_j\}$,
    \begin{align*}
        \bigl|\rho(X_i, X_j \mid X_S)\bigr| > \lambda
        \;\iff\;
        X_i \not\perp_{\mathrm{d\text{-}sep}} X_j \mid X_S,
    \end{align*}
    that is, every partial correlation corresponding to a d-connection is bounded away from zero by $\lambda$.
\end{definition}

\textbf{Remark:} \citet{zhang2012strong} and \citet{uhler2013geometry} state $\lambda$-strong faithfulness
for linear Gaussian models. We state it for arbitrary distributions with finite second
moments, since the partial correlations in Definition~\ref{def:strong-faithful} are
well-defined whenever those exist, and we consider it beyond that case. 
Covariance-faithfulness (Definition~\ref{def:cov-faithful}) is the excluded limiting case $\lambda = 0$.

\begin{definition}[Varsortability and $R^2$-sortability]\label{def:sortability}
    Let $\mathbb{P}$ be a distribution over $V = \{X_1, \dots, X_p\}$ that is Markov to a DAG
    $\mathcal{G}$ with at least one edge, and let $\tau$ assign a value $\tau_i$ to each node.
    The \emph{$\tau$-sortability} $v_\tau(\mathbb{P}, \mathcal{G}) \in [0,1]$ is the fraction of
    triples $(X_i, X_j, k)$ for which $\mathcal{G}$ contains a directed path of length $k$ from
    $X_i$ to $X_j$ and $\tau_i < \tau_j$, counting ties as one half
    \citep{reisach2021beware, reisach2023scale}.
    \emph{Varsortability} takes $\tau_i = \mathrm{Var}(X_i)$, and \emph{$R^2$-sortability} takes
    $\tau_i = R^2_i$, the fraction of the variance of $X_i$ explained by the remaining variables,
    which in the linear case equals $1 - \bigl(\mathrm{Var}(X_i)\,(\Sigma^{-1})_{ii}\bigr)^{-1}$
    for $\Sigma$ the covariance matrix of $\mathbb{P}$.
    Either measure equals $1$ iff every node's criterion value strictly exceeds that of all its
    ancestors.
\end{definition}

\section{Extended Discussion of Examples}\label{app:examples}

Here, we provide the full details of Examples~\ref{ex:one} and \ref{ex:two}.
Together, they showcase that without further assumptions there is no principled reason to prefer faithfulness over covariance-faithfulness.
Hence, they exemplify the implicit standards in the causal discovery literature of how we decide which identifiability assumptions count as fundamental.

\subsection{Example 1: Covariance-faithfulness does not imply faithfulness}

Consider the data-generating process described by the following SCM:
\begin{align*}
    X_1 &= N_1 \\
    X_2 &= -3 + 3 \cdot \mathbf{1}\{X_1=2\} + N_2 \\
    X_3 &= -3 + \mathbf{1}\{X_1 > 0\} + \operatorname{sign}(X_2) \cdot N_3
\end{align*}
where $N_1\sim \mathrm{Unif}(\{0,1,2\}), N_2,N_3\sim \mathrm{Unif}(\{-1,1\})$
are jointly independent
and $\mathbf{1}\{\cdot\}$ denotes the indicator function.
This SCM induces an observational joint distribution $\mathbb{P}$ over $12$ configurations,
each occurring with equal probability.
The $12$ equally likely configurations are
\begin{center}
\begin{tabular}{r|rrrrrrrrrrrr}
$X_1$ & $0$ & $0$ & $0$ & $0$ & $1$ & $1$ & $1$ & $1$ & $2$ & $2$ & $2$ & $2$ \\
$X_2$ & $-4$ & $-4$ & $-2$ & $-2$ & $-4$ & $-4$ & $-2$ & $-2$ & $-1$ & $-1$ & $1$ & $1$ \\
$X_3$ & $-2$ & $-4$ & $-2$ & $-4$ & $-1$ & $-3$ & $-1$ & $-3$ & $-1$ & $-3$ & $-3$ & $-1$
\end{tabular}
\end{center}
The SCM is structurally minimal (Remark 6.6 in \citealp{peters2017eleci})
and the causal graph $\mathcal{G}_0$ therefore is the fully connected triangle
\begin{tikzpicture}[baseline=(B.base), overlay=false]
\node (A) {$X_1$};
\node (B) [right=0.9em of A] {$X_2$};
\node (C) [right=0.9em of B] {$X_3$};
\draw[->, line width=0.3pt] (A) -- (B);
\draw[->, line width=0.3pt] (B) -- (C);
\draw[->, bend left=18, line width=0.3pt] (A) to (C);
\end{tikzpicture}\!.
Direct computation yields the covariance and precision matrices
    \begin{align*}
        \Sigma = 
        \frac{1}{9}
        \begin{pmatrix}
            6 & 9 & 3 \\
            9 & 27 & 3 \\
            3 & 3 & 11
        \end{pmatrix}, \qquad
        \Sigma^{-1} = 
        \frac{1}{28}
        \begin{pmatrix}
            96 & -30 & -18 \\
            -30 & 19 & 3 \\
            -18 & 3 & 27
        \end{pmatrix}.
    \end{align*}
All entries of $\Sigma$ are nonzero, and all entries of $\Sigma^{-1}$ are nonzero.
Therefore, $\mathbb{P}$ is covariance-faithful to $\mathcal{G}_0$.
Yet, $\mathbb{P}$ is not faithful to $\mathcal{G}_0$: enumeration of the $12$ configurations shows that $X_2 \indep X_3 \mid X_1$ and that this is the only (conditional) independence among $X_1, X_2, X_3$.
Specifically, conditional on $X_1 \in \{0,1\}$, $\operatorname{sign}(X_2) = -1$ deterministically, so $X_3 = -3 + \mathbf{1}\{X_1>0\} - N_3$ depends only on $X_1$ and the independent noise $N_3$;
conditional on $X_1 = 2$, $X_2 = N_2$ and $X_3 = -2 + \operatorname{sign}(X_2) N_3$, 
so $X_3 \mid X_1=2, X_2=x_2$ is uniform on $\{-3,-1\}$ regardless of the value $x_2$.
The graph $\mathcal{G}_0$, however, contains the edge $X_2 \to X_3$, so no conditioning set
d-separates $X_2$ from $X_3$ and the conditional independence 
$X_2 \indep X_3 \mid X_1$ 
has no counterpart in $\mathcal{G}_0$.
By contrast, $\mathbb{P}$ is faithful, for example, to the chain $X_2 \to X_1 \to X_3$, whose only d-separation statement is $X_2 \indep X_3 \mid X_1$.

\subsection{Example 2: Faithfulness does not imply covariance-faithfulness}

Consider the data-generating process described by the following SCM:
\begin{align*}
    X_1 &= N_1 \\
    X_2 &= X_1 + N_2 \\
    X_3 &= 2X_1 - X_2 + N_3
\end{align*}
where $N_1, N_2, N_3$ are jointly independent with zero mean, unit variance, and non-Gaussian distribution with finite fourth moment.
Substituting yields the reduced form $X_2 = N_1 + N_2$ and $X_3 = N_1 - N_2 + N_3$. 
It is structurally minimal and the causal graph $\mathcal{G}_0$ is the fully connected triangle
\begin{tikzpicture}[baseline=(B.base), overlay=false]
\node (A) {$X_1$};
\node (B) [right=0.9em of A] {$X_2$};
\node (C) [right=0.9em of B] {$X_3$};
\draw[->, line width=0.3pt] (A) -- (B);
\draw[->, line width=0.3pt] (B) -- (C);
\draw[->, bend left=18, line width=0.3pt] (A) to (C);
\end{tikzpicture}\!.
Direct computation yields the covariance and precision matrices
    \begin{align*}
        \Sigma = 
        \begin{pmatrix}
            1 & 1 & 1 \\
            1 & 2 & 0 \\
            1 & 0 & 3
        \end{pmatrix}, \qquad
        \Sigma^{-1} = 
        \begin{pmatrix}
            6 & -3 & -2 \\
            -3 & 2 & 1 \\
            -2 & 1 & 1
        \end{pmatrix}.
    \end{align*}
Thus, covariance-faithfulness to $\mathcal{G}_0$ is not fulfilled, since $\Sigma_{23} = 0$ despite $X_2$ and $X_3$ being adjacent in $\mathcal{G}_0$.
Yet, $\mathbb{P}$ is faithful to $\mathcal{G}_0$, as we verify by checking that no (conditional) independence holds among $X_1, X_2, X_3$.
The marginal dependences $X_1 \nindep X_2$ and $X_1 \nindep X_3$ follow from $\Sigma_{12} = \Sigma_{13} = 1 \neq 0$.
Despite vanishing covariance $\Sigma_{23}=0$, we have $X_2\nindep X_3$:
Since $X_2 = N_1 + N_2$ and $X_3 = N_1 - N_2 + N_3$ are linear combinations of $N_1$ and $N_2$ with nonzero coefficients, the Darmois--Skitovich theorem implies that $X_2 \indep X_3$ would force $N_1$ and
$N_2$ to be Gaussian. However, by assumption they are non-Gaussian, so $X_2 \nindep X_3$.
For the conditional dependences, note that $X_3 \mid X_1 = x_1, X_2 = x_2 \sim 2x_1 - x_2 + N_3$ depends on both $x_1$ and $x_2$.
Since $N_2$ is non-degenerate, conditioning on $X_2$ alone leaves $X_1$ random, so the conditional law of $X_3 \mid X_2$ still depends on $X_1$, giving $X_3 \nindep X_1 \mid X_2$; the symmetric argument with $X_1$ and $X_2$ exchanged gives $X_3 \nindep X_2 \mid X_1$.
Finally, the joint conditional density of $(X_1, X_2)$ given $X_3 = x_3$ satisfies
\(
    p(x_1, x_2 \mid x_3) \propto p_{N_1}(x_1)\, p_{N_2}(x_2 - x_1)\, p_{N_3}(x_3 - 2x_1 + x_2),
\)
which does not factorize into a function of $x_1$ alone times a function of $x_2$ alone, since $p_{N_3}(x_3 - 2x_1 + x_2)$ couples $x_1$ and $x_2$; hence $X_1 \nindep X_2 \mid X_3$.

\section{Additional Experimental Details}\label{app:exps}

Here, we give an overview of the experimental details and provide additional figures. 

\subsection{Data Generation}

\begin{enumerate}
    \item \textbf{Graph generation}: We consider ER graphs and chains without colliders.
    To sample a chain without colliders, we first build a chain skeleton and shuffle the node indices. We then randomly select a source node and orient all edges away from it.
    \item \textbf{Edge weights}: Unless specified otherwise, we draw edge coefficients uniformly from $\pm [0.25, 1.0]$.
    \item \textbf{Noise distributions}: 
    The noise is drawn either from a generalized normal distribution, where the degree of Gaussianity is controlled by the $\beta$-parameter (1.0 corresponds to the Laplace distribution, 2.0 to the Gaussian distribution), or from a zero-centered uniform distribution (with the bounds set to meet the desired variance).
    Unless specified otherwise, we draw the variances of the noise distributions uniformly from $[0.5, 2.0]$.
    \item \textbf{MEC size computation}: We first sample an ER-DAG, then obtain its CPDAG by calling \texttt{dag2cpdag} in the \texttt{causal-learn} Python package \citep{zheng2024causal}, and compute the MEC size using the \texttt{CliquePicking} Python package available at \url{https://github.com/mwien/CliquePicking}.
\end{enumerate}

\subsection{Algorithms}

\begin{enumerate}
    \item \textbf{FLOP}: We use the implementation from the Python package available at \url{https://github.com/CausalDisco/flopsearch}.
    Unless specified otherwise, we set the number of restarts to 20 and the BIC penalty to 2.0, following the original paper.
    FLOP outputs a CPDAG, but we use it as a DAG-learner.
    To sample a DAG from the MEC, we randomly select an undirected edge and orient it in a random direction.
    We then apply Meek's rules to obtain a CPDAG again, and repeat this process until all edges are oriented.
    \item \textbf{DirectLiNGAM}: We use the implementation from the official Python package of DirectLiNGAM \citep{ikeuchi2023pythonlingam}.
    We use the default parameters throughout.
    \item \textbf{VarSortRegress}: Compared to the existing implementation at \url{https://github.com/CausalDisco/CausalDisco/}, we standardize the data before parent selection.
    We did not observe any differences between the results of the existing implementation and ours.
\end{enumerate}

\subsection{Figures}\label{app:additionalfigures}

\begin{figure}[ht]
    \centering
    \includegraphics[width=\linewidth]{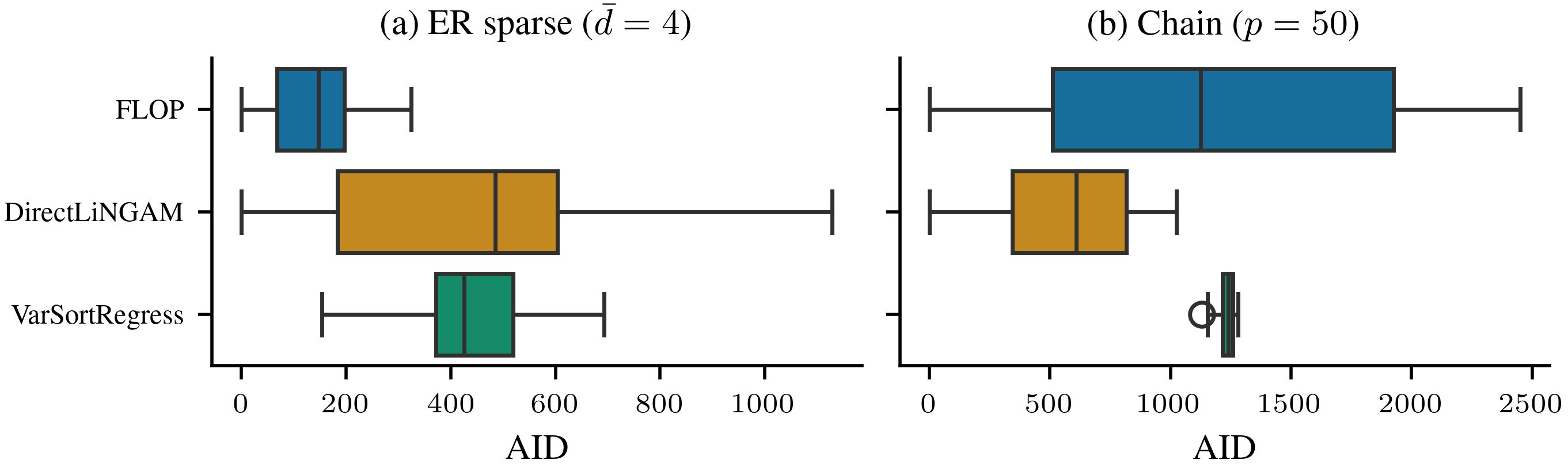}
    \caption{As Figure~\ref{fig:MEC+type}(b) and (c), but reporting ancestor AID instead of SHD.}
    \label{fig:graphtype-aid}
\end{figure}

\begin{figure}[ht]
    \centering
    \includegraphics[width=\linewidth]{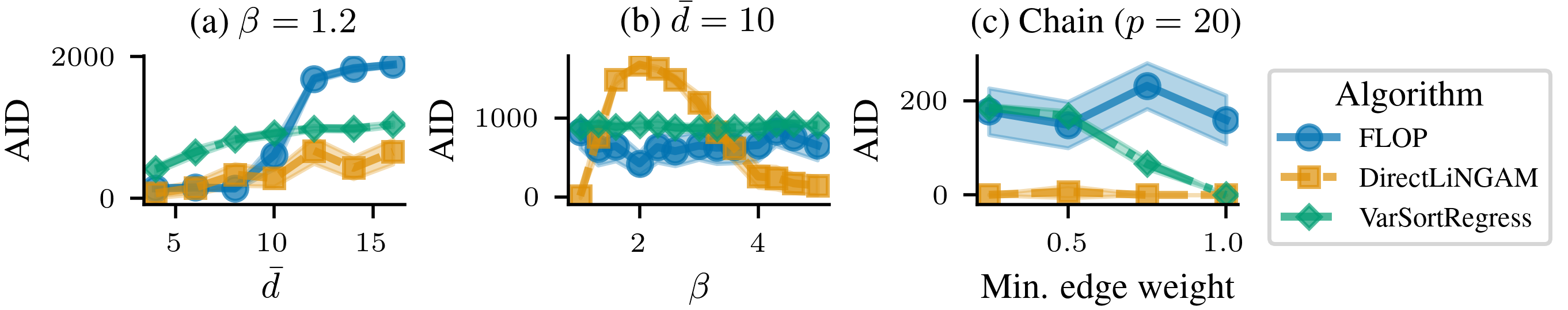}
    \caption{As Figure~\ref{fig:axis}, but reporting ancestor AID instead of SHD.}
    \label{fig:degradationaid}
\end{figure}

\begin{figure}[ht]
    \centering
    \includegraphics[width=\linewidth]{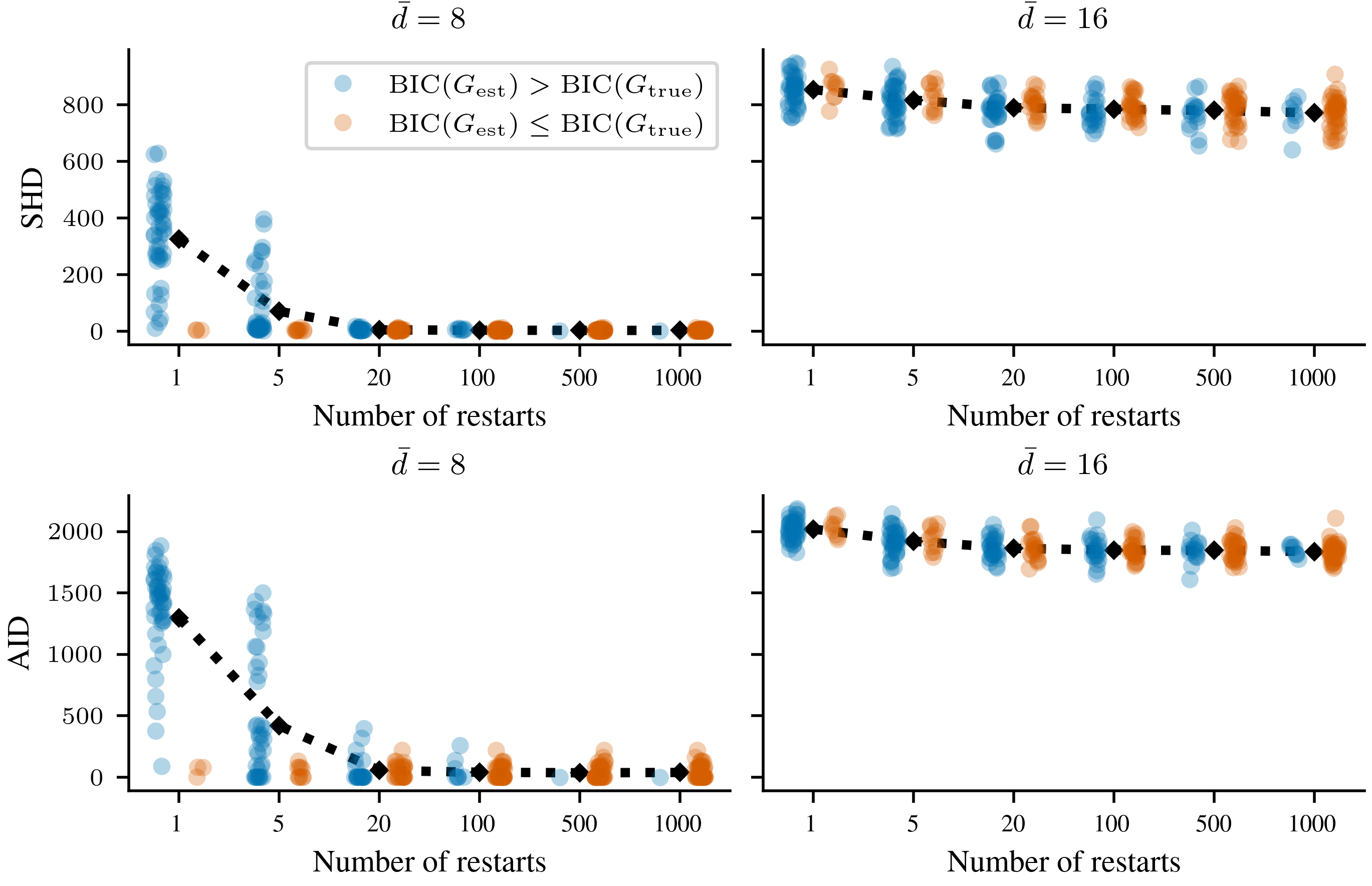}
    \caption{
    SHD and ancestor AID vs.\ number of restarts in FLOP, on linear ANMs with $p=50$ nodes,
    $n=1000$ samples, and uniform noise; results across 50 sampled graphs per setting, with
    average degree $\bar{d}=8$ (left) and $\bar{d}=16$ (right).
    Lower BIC is better, so points with $\mathrm{BIC}(G_{\mathrm{est}}) \leq \mathrm{BIC}(G_{\mathrm{true}})$ mark runs in which the search found a graph scoring at least as well as the true graph.
    With more restarts FLOP increasingly does so, but for $\bar{d}=16$ this does not translate to recovering the true graph: the search succeeds while the graph remains wrong.
    }
    \label{fig:bic}
\end{figure}

\begin{figure}[ht]
    \centering
    \includegraphics[width=\linewidth]{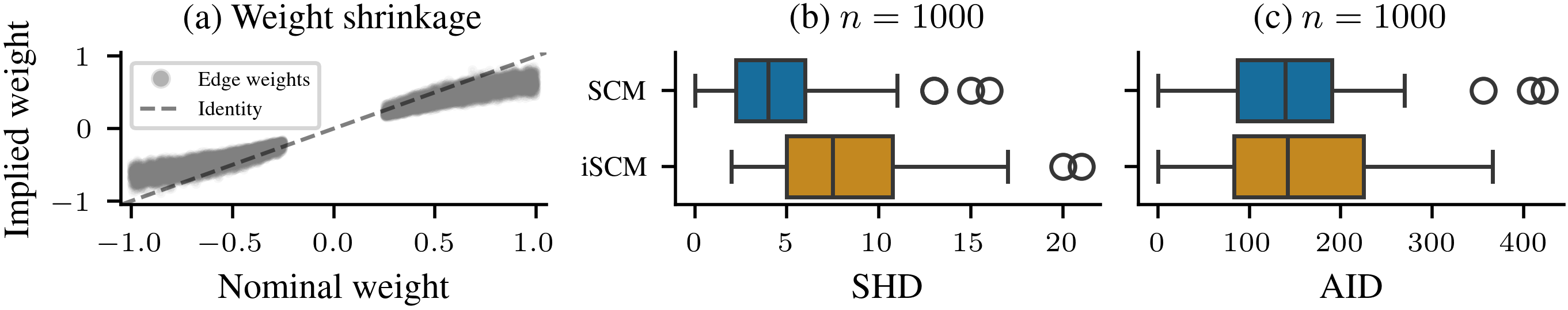}
    \caption{%
    \textit{(a)}: The implied edge weights of an iSCM tend to be smaller than the originally drawn weights, where implied weights are computed via Algorithm~2 in \citet{ormaniec25iscm}. \textit{(b) and (c)}: FLOP exhibits a slightly worse performance on the iSCMs compared to standard ER graphs.
    Both use $p=50$ nodes, $\bar{d}=8$, and Gaussian noise.
}
    \label{fig:iscm}
\end{figure}

\begin{figure}[ht]
    \centering
    \includegraphics[width=\linewidth]{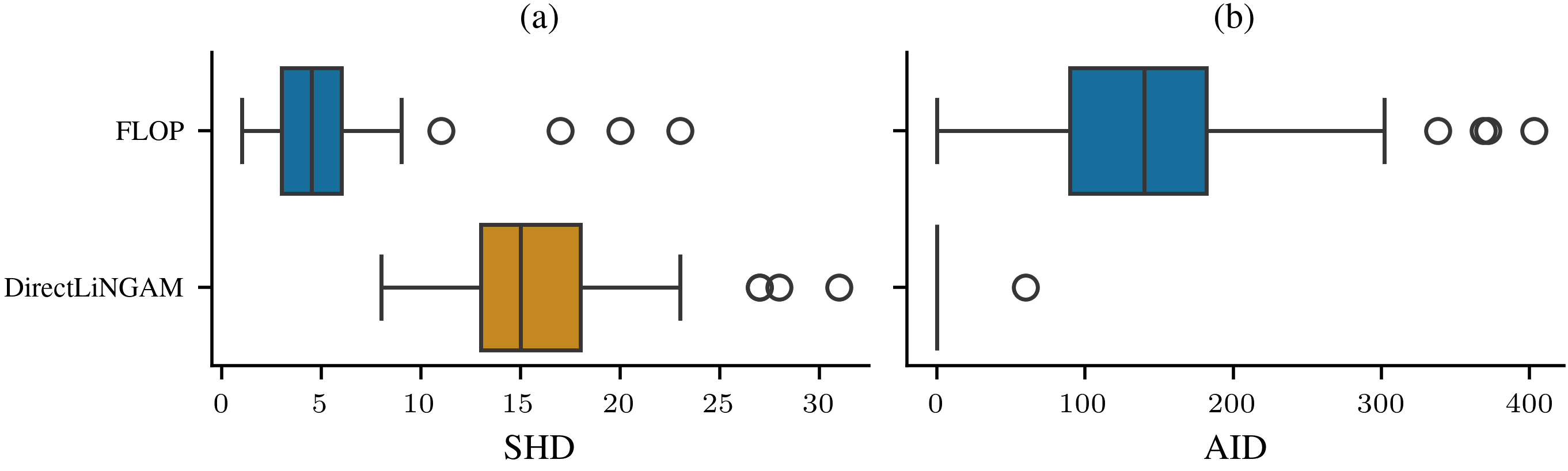}
    \caption{FLOP vs.\ DirectLiNGAM across two different metrics, SHD and ancestor AID, on linear ANMs with $p=50$ node ER graphs with $\bar{d}=8$, $n=1000$ samples, and uniform noise (variances uniformly sampled from $[0.5, 2.0]$); results across 50 sampled graphs. FLOP outperforms DirectLiNGAM with respect to SHD, but DirectLiNGAM outperforms FLOP with respect to ancestor AID, illustrating that the choice of metric influences results. VarSortRegress is omitted because its performance is not informative here.}
    \label{fig:metric}
\end{figure}

\end{document}